\documentclass[11pt,letterpaper]{article}
\usepackage[letterpaper]{geometry}
\usepackage{acl2012}
\usepackage{times}
\usepackage{latexsym}
\usepackage{amsmath,amsfonts,amssymb}
\usepackage{multirow}
\usepackage{graphicx}
\usepackage{url}
\usepackage{color}
\makeatletter
\newcommand{\@BIBLABEL}{\@emptybiblabel}
\newcommand{\@emptybiblabel}[1]{}
\makeatother
\usepackage[hidelinks]{hyperref}
\DeclareMathOperator*{\argmax}{arg\,max}
\title{Low-Rank RNN Adaptation for Context-Aware Language Modeling}
\author{Aaron Jaech \and Mari Ostendorf \\
University of Washington \\
Seattle, WA \\ 
{\tt \{ajaech,ostendor\}@uw.edu}
}
\date{}
\begin{document}

\maketitle

\begin{abstract}
A context-aware language model uses location, user and/or domain metadata (context) to adapt its predictions. In neural language models, context information is typically represented as an embedding and it is given to the RNN as an additional input, which has been shown to be useful in many applications. We introduce a more powerful mechanism for using context to adapt an RNN by letting the context vector control a low-rank transformation of the recurrent layer weight matrix. Experiments show that allowing a greater fraction of the model parameters to be adjusted has benefits in terms of perplexity and classification for several different types of context.
\end{abstract}

\section{Introduction}

In many language modeling applications, the speech or text is associated with some metadata or contextual information. For example, in speech recognition, if a user is speaking to a personal assistant then the system might know the time of day or the identity of the task that the user is trying to accomplish. If the user takes a picture of a sign to translate it with their smart phone, the system would have contextual information related to the geographic location and the user's preferred language. The context-aware language model targets these types of applications with a model that can adapt its predictions based on the provided contextual information.

There has been much work on using context information to adapt language models. Here, we are interested in contexts described by metadata (vs.\ word history or related documents) and in neural network approaches due to their flexibility for representing diverse types of contexts. Specifically, we focus on recurrent neural networks (RNNs) due to their widespread use. 

The standard approach to adapt an RNN language model is to concatenate the context representation with the word embedding at the input to the RNN \cite{mikolov2012context}. Optionally, the context embedding is also concatenated with the output from the recurrent layer to adapt the softmax layer. This basic strategy has been adopted for various types of adaptation such as for LM personalization \cite{wen2013recurrent,li2016persona}, adapting to television show genres \cite{chen2015recurrent}, adapting to long range dependencies in a document \cite{Ji2015DocumentCL}, etc.

We propose a more powerful mechanism for using a context vector, which we call the FactorCell. Rather than simply using context as an additional input, it is used to control a factored (low-rank) transformation of the recurrent layer weight matrix. The motivation is that allowing a greater fraction of the model parameters to be adjusted in response to the input context will produce a model that is more adaptable and responsive to that context.

We evaluate the resulting models in terms of context-dependent perplexity and context classification accuracy on six tasks reflecting different types of context variables, comparing to baselines that represent the most popular methods for using context in neural models. 
We choose tasks where context is specified by metadata, rather than text samples as used in many prior studies.
The combination of experiments on a variety of data sources provides strong evidence for the utility of the FactorCell model, but the results show that it can be useful to consider more than just perplexity in training a language model.

The remainder proceeds as follows. In Section \ref{sec:model}, we introduce the FactorCell model and show how it differs mathematically from alternative approaches. Next, Section \ref{sec:data} describes the six datasets used to probe the performance of different models. Experiments and analyses contrasting perplexity and classification results for a variety of context variables are provided in Section \ref{sec:experiments}, demonstrating consistent improvements in both criteria for the FactorCell model but also confirming that perplexity is not correlated with classification performance for all models. Analyses 
explore the effectiveness of the model  for characterizing high-dimensional context spaces. The model is compared to related work in Section~\ref{sec:prior}. Finally, Section~\ref{sec:concl} summarizes contributions and open questions.

\section{Context-Aware RNN}
\label{sec:model}

Our model uses adaptation in both the recurrent layer and in the bias vector of the output layer. In this section we describe how we represent context as an embedding 
and methods for adapting the recurrent layer and the softmax layer, showing that our proposed model is a generalization of prior methods. 
The novelty of our model is that instead of using context as an additional input to the model, it uses the context information to transform the weights of the recurrent layer. This is accomplished using a low-rank decomposition in order to control the extent of parameter sharing between contexts,
which is important for handling high-dimensional, sparse contexts.

\subsection{Context representation}

We assume the availability of contextual information (metadata or other side information) that is represented as a set of
context variables $f_{1:n}=f_1, f_2, \dots f_n$, from which we produce a $k$-dimensional representation in the form of an embedding, $c \in \mathbb{R}^k$. Each of the context variables, $f_i$, represents some type of information or metadata about the sequence and can be either categorical or numerical. The embeddings can either be learned off-line using a topic model \cite{mikolov2012context} or end-to-end as part of the adapted LM \cite{TangContextAware}.  Here, we use end-to-end learning, where the context embedding is the output of a feed-forward network with a ReLU activation function.
The resulting embedding, $c$, is used for adapting both the recurrent layer and the output layer of the RNN. 

\subsection{Adapting the recurrent layer}

The basic operation of the recurrent layer is to use a matrix $\mathbf{W}$ to transform the concatenation of a word embedding, $w_t \in \mathbb{R}^e$, with the hidden state from the previous time step, $h_{t-1} \in \mathbb{R}^d$, and produce a new hidden state, $h_t$, as given by Equation \ref{eq:rnn}:  
\begin{align}
\label{eq:rnn}
\begin{split}
h_t &= \sigma(\mathbf{W}_1 w_t + \mathbf{W}_2 h_{t-1} + b) \\
    &= \sigma(\mathbf{W}[w_t, h_{t-1}] + b).
\end{split}
\end{align}
The size of $\mathbf{W}$ is $d\times (e+d)$. For simplicity, our equations assume a simple RNN. Appendix \ref{sec:lstm_equations} shows how the equations can be adjusted to work with an LSTM. 

The standard approach to recurrent layer adaptation is to include (via concatenation) the context embedding as an additional input to the recurrent layer \cite{mikolov2012context}. When the context embedding is constant across the whole sequence, it is easy to show that this concatenation is equivalent to using a context-dependent bias at the recurrent layer:  
\begin{align}
\label{eq:ConcatCell}
\begin{split}
  h_t &= \sigma( \hat{\mathbf{W}}[w_t, h_{t-1}, c] + b) \\
      &= \sigma( \mathbf{W}[w_t, h_{t-1}] + \mathbf{V}c + b) \\
      &= \sigma( \mathbf{W}[w_t, h_{t-1}] + b'),
\end{split}
\end{align}
where 
$\hat{\mathbf{W}} = [ \mathbf{W} \  \mathbf{V} ]$ 
and $b' = \mathbf{V}c + b$ is the context-dependent bias, formed by adding a linear projection of the context embedding. We refer to this adaptation approach as the ConcatCell model. 

Our proposed model extends the ConcatCell by using a context-dependent weight matrix
$\mbox{$\mathbf{W}' = \mathbf{W} + \mathbf{A}$}$, in place of the generic weight matrix $\mathbf{W}$. (We refer to $\mathbf{W}$ as generic because it is shared across all context settings.) 
The adaptation matrix, $\mathbf{A}$, is generated by taking the product of the context embedding vector against a set of left and right basis tensors to produce a rank $r$ matrix. 
The left and right adaptation basis tensors are given as  $\mathbf{Z}_L \in \mathbb{R}^{k \times (e+d) \times r}$ and $\mathbf{Z}_R \in \mathbb{R}^{r \times d \times k}$. The two bases tensors together can be thought of as holding $k$ different rank $r$ matrices, $A_j=\mathbf{Z}_{L,j}\mathbf{Z}_{R,j}$, each the size of $\mathbf{W}$. By taking the product between $c$ and the corresponding tensor modes of $\mathbf{Z}_L$ and $\mathbf{Z}_R$ (using $\times_i$ to denote the mode-$i$ tensor product, i.e., the product with the $i$-th dimension of the tensor), the context determines the weighted combination of the $k$ matrices:
\begin{equation}
\label{eq:w_prime}
  A=(c \times_1 \mathbf{Z}_L)(\mathbf{Z}_R \times_3 c^\intercal) .  
\end{equation}
The number of degrees of freedom of $A$ is controlled by the dimension $k$ of the context vector and the rank $r$ of the $k$ weight matrices. The rank is treated as a hyperparameter and controls the extent to which the model relies on the generic weight matrix $\mathbf{W}$ versus behaves in a more context-specific manner.

We call this model the FactorCell because the weight matrix has been adapted by adding a factored component. The ConcatCell model is a special case of the FactorCell where $\mathbf{Z}_L$ and $\mathbf{Z}_R$ are set to zero. In summary, the proposed model is given by:
\begin{align}
\label{eq:FactorCell}
\begin{split}
  h_t &= \sigma( \mathbf{W}'[w_t, h_{t-1}] + b') \\
\mathbf{W}'  &=  \mathbf{W} + (c \times_1 \mathbf{Z}_L)(\mathbf{Z}_R \times_3 c)\\
 b'  &=  \mathbf{V}c + b.
\end{split}
\end{align}

If the context is known in advance, $\mathbf{W}'$ can be precomputed, in which case applying the RNN at test time requires no more computation than using an unadapted RNN of the same size. This means that for a fixed sized recurrent layer, the FactorCell model can have many more parameters than the ConcatCell model but hardly any increase in computational cost.

\subsection{Adapting the Softmax Bias}
\label{sec:softmaxbias}

The last layer of the model predicts the probability of the next symbol in the sequence using the output from the recurrent layer using the softmax function to create a normalized probability distribution. The output probabilities are given by 
\begin{equation}
y_t = \mathrm{softmax}(\mathbf{E}\mathbf{L}h_t + b_{out}),
\end{equation}
where $\mathbf{E}\in\mathbb{R}^{|V|\times e}$ is the matrix of word embeddings, $\mathbf{L}\in\mathbb{R}^{e\times d}$ is a linear projection to match the dimension of the recurrent layer (when $e\ne d$), and $b_{out} \in \mathbb{R}^{|V|}$ is the softmax bias vector. 
We tie the word embeddings in the input layer with the ones in the output layer \cite{press2016using,inan2016tying}. 

If $s_j$ is the indicator row vector for the $j$th word in the vocabulary then $p(w_t|w_{1:t-1}) = s_t y_t$ and $\log p(w_{1:T}) = \sum_t \log s_{w_t} y_t$.

Adapting the softmax bias alters the unigram distribution. There are two ways to accomplish this. When the values that the context can take are categorical with low cardinality then context-dependent softmax bias vectors can be learned directly. This is equivalent to replacing $c$ with a one-hot vector. Otherwise, a projection of the context embedding, $\mathbf{Q}c$ where $\mathbf{Q} \in \mathbb{R}^{|V| \times k}$, can be used to adapt the bias vector as in 
\begin{equation}
\label{eqn:adaptSoftmax}
y_t = \mathrm{softmax}(\mathbf{E}\mathbf{L}h_t + \mathbf{Q}c + b_{out}).
\end{equation}
The projection can be thought of as a low-rank approximation to using the one-hot context vector. Both strategies are explored,
depending on the nature of the original context space.

As noted in Section~\ref{sec:prior}, adaptation of the softmax bias has been used in other studies. As we show in the experimental work, it is useful for representing phenomena where unigram statistics are important.

\section{Data}
\label{sec:data}

\begin{table*}[h]
\centering
\begin{tabular}{crrrrrc}
\textbf{Name} & \textbf{Train} & \textbf{Dev} & \textbf{Test} & \textbf{Vocab} & \textbf{Docs.} & \textbf{Context} \\ \hline
AGNews & 4.6M & 0.2M & 0.3M & 54,492 & 115K & 4 Newspaper sections \\
DBPedia & 28.7M & 0.3M & 3.6M & 84,341 & 555K & 14 Entity categories \\
TripAdvisor & 127.2M & 2.6M & 2.6M & 88,347 & 843K & 3.5K Hotels/5 Sentiment \\
Yelp & 91.5M & 0.7M & 7.1M & 57,794 & 645K & 5 Sentiment \\
EuroTwitter$^*$ & 5.3M &  0.8M & 1.0M & 194 & 80K & 9 Languages \\
GeoTwitter$^*$ & 51.7M & 2.2M & 2.2M & 203 & 604K & Latitude \& Longitude
\end{tabular}
\caption{Dataset statistics: Dataset size in words (* or characters) of Train, Dev and Test sets, vocabulary size, number of training documents, and context variables.}
\label{table:data}
\end{table*}

Our experiments make use of six datasets: four targeting word-level sequences, and two targeting character sequences. The character studies are motivated by the growing interest in character-level models in both speech recognition and machine translation \cite{hannun2014deep,chung2016character}. By using multiple datasets with different types of context, we hope to learn more about what makes a dataset amenable to adaptation. The datasets range in size from over 100 million words of training data to 5 million characters of training data for the smallest one.
When using a word-based vocabulary, we preprocess the data by lowercasing, tokenizing and removing most punctuation. We also truncate sentences to be shorter than a maximum length of 60 words for AGNews and DBPedia and 150 to 200 tokens for the remaining datasets. 
Summary information is provided in Table~\ref{table:data}, including the training, development, and test data sizes in terms of number of tokens, vocabulary size, number of training documents (i.e. context samples), and the context variables ($f_{1:n}$). The largest dataset, TripAdvisor, has over 800 thousand hotel review documents, which adds up to over 125 million words of training data. 

The first three datasets (AGNews, DBPedia, and Yelp) have previously been used for text classification \cite{zhang2015character}.
These consist of newspaper headlines, encyclopedia entries, and restaurant and business reviews, respectively. The context variables associated with these correspond to the newspaper section (world, sports, business, sci \& tech) for each headline, the page category on DBPedia (out of 14 options such as actor, athlete, building, etc.), and the star rating on Yelp (from one to five).  For AgNews, DBPedia, and Yelp we use the same test data as in previous work. Our fourth dataset, from TripAdvisor, was previously used for language modeling and consists of two relevant context variables: an identifier for the hotel and a sentiment score from one to five stars \cite{TangContextAware}. Some of the reviews are written in French or German but most are in English. There are 4,333 different hotels but we group all the ones that do not occur at least 50 times in the training data into a single entity, leaving us with around 3,500. These four datasets use word-based vocabularies. 

We also experiment on two Twitter datasets: EuroTwitter and GeoTwitter. EuroTwitter consists of 80 thousand Tweets labeled with one of nine languages: (English, Spanish, Galician, Catalan, Basque, Portuguese, French, German, and Italian). The corpus was created by combining portions of multiple published datasets for language identification including Twitter70 \cite{jaech2016hierarchical}, TweetLID \cite{zubiaga2014overview}, and the monolingual portion of Tweets from a code-switching detection workshop \cite{molina2016overview}. The GeoTwitter data contains Tweets with latitude and longitude information from England, Spain, and the United States.\footnote{Data was accessed from http://followthehashtag.com.} The latitude and longitude coordinates are given as numerical inputs. This is different from the other five datasets that all use categorical context variables.

\section{Experiments with Different Contexts}
\label{sec:experiments}

The goal of our experiments is to show that the FactorCell model can deliver improved performance over current approaches for multiple language model applications and a variety of types of contexts. Specifically, results are reported for context-conditioned perplexity and generative model text classification accuracy, using contexts that capture a range of phenomena and dimensionalities.

Test set perplexity 
is the most widely accepted method for evaluating language models, both for use in recognition/translation applications and generation. It has the advantage that it is easy to measure and is widely used as a criteria for model fit, but the limitation that it is not directly matched to most tasks that language models are directly used for. 
Text classification using the model in a generative classifier is a simple application of Bayes rule:
\begin{equation}
\label{eq:bayes}
    \hat \omega = \argmax_\omega p(w_{1:T} | \omega) p(\omega)
\end{equation}
where $w_{1:T}$ is the text sequence, $p(\omega)$ is the class prior, which we assume to be uniform.
Classification accuracy provides additional information about the power of a model, even if it is not being designed explicitly for text classification. Further, it allows us to be able to directly compare our model performance against previously published text classification benchmarks.

Note that the use of classification accuracy for evaluation here involves counting errors associated with applying the generative model to independent test samples. This differs from the accuracy criterion used for evaluating context-sensitive language models for text generation based on a separate discriminative classifier trained on generated text \cite{Ficler2017ControllingLS,Hu2017ControllableTG}. We discuss this further in Section~\ref{sec:prior}.

The experiments compare the FactorCell model (equations \ref{eq:FactorCell} and \ref{eqn:adaptSoftmax}) to two popular alternatives, which we refer to as ConcatCell (equations \ref{eq:ConcatCell} and \ref{eqn:adaptSoftmax}) and SoftmaxBias (equation \ref{eqn:adaptSoftmax}). As noted earlier, the SoftmaxBias method is a simplification of the ConcatCell model, which is in turn a simplification of the FactorCell model. The SoftmaxBias method impacts only the output layer and thus only unigram statistics. 
Since bag-of-word models provide strong baselines in many text classification tasks, we hypothesize that the SoftmaxBias model will capture much of the relative improvement over the unadapted model for word-based tasks. However, in small vocabulary character-based models, the unigram distribution is unlikely to carry much information about the context, so adapting the recurrent layer should become more important in character-level models. 
We expect that performance gains will be greatest for the FactorCell model for sources that have sufficient structure and data to support learning the extra degrees of freedom.

Another possible baseline would use models independently trained on the subset of data for each context. This is the ``independent component'' case in \cite{Yogatama2017GenerativeAD}. This will fail when a context variable takes on many values (or continuous values) or when training data is limited, because it makes poor use of the training data, as shown in that study. While we do have some datasets where this approach is plausible, we feel that its limitations have been clearly established.

\subsection{Implementation Details}
The RNN variant that we use is an LSTM with coupled input and forget gates \cite{melis2017state}.  The different model variants are implemented\footnote{Code available at http://github.com/ajaech/calm.} using the Tensorflow library. 
The model is trained with the standard negative log likelihood loss function, i.e.\ minimizing cross entropy.
Dropout is used as a regularizer in the recurrent connections as described in \newcite{semeniuta2016recurrent}. Training is done using the Adam optimizer with a learning rate of $0.001$. For the models with word-based vocabularies, a sampled softmax loss is used with a unigram proposal distribution and sampling 150 words at each time-step \cite{jean2014using}. The classification experiments use a sampled softmax loss with a sample size of 8,000 words. This is an order of magnitude faster to compute with a minimal effect on accuracy.

\begin{table*}[]
\centering
\begin{tabular}{crrrrrr}
\textbf{}  & \textbf{AgNews} & \textbf{DBPedia} & \textbf{EuroTwitter} & \textbf{GeoTwitter} & \textbf{TripAdvisor} & \textbf{Yelp} \\ \hline
Word Embed & 150      & 114-120  & 35-40       & 42-50      & 100         & 200      \\
LSTM dim   & 110      & 167-180  & 250         & 250        & 200         & 200      \\
Steps      & 4.1-5.5K & 7.5-8.0K & 6.0-8.0K    & 6.0-11.1K  & 8.4-9.9K    & 7.2-8.8K \\
Dropout    & 0.5      & 1.00 & 0.95-1.00   & 0.99-1.00      & 0.97-1.00   & 1.00     \\
Ctx. Embed & 2        & 12      & 3-5         & 8-24       & 20-30       & 2-3      \\
Rank       & 12       & 19        & 2           & 20         & 12          & 9       
\end{tabular}
\caption{Selected hyperparameters for each dataset. When a range is listed it means that a different values were selected for the FactorCell, ConcatCell, SoftmaxBias or Unadapted models.}
\label{table:hyperparams2}
\end{table*}

Hyperparameter tuning was done based on minimizing perplexity on the development set and using a random search. 
Hyperparameters included word embedding size $e$, recurrent state size $d$, context embedding size $k$, and weight adaptation matrix rank $r$, the number of training steps, recurrent dropout probability, and random initialization seed. The selected hyperparameter values are listed in Table \ref{table:hyperparams2}
For any fixed LSTM size, the FactorCell has a higher count of learned parameters compared to the ConcatCell. However, during evaluation both models use approximately the same number of floating-point operations because $\mathbf{W}'$ only needs to be computed once per sentence. Because of this, we believe limiting the recurrent layer cell size is a fair way to compare between the FactorCell and the ConcatCell.

\subsection{Word-based Models}
\label{sec:word_based_models}

\begin{table*}[h!]
\centering
\begin{tabular}{c|rr|rr|rr|rr}
 & \multicolumn{2}{c|}{\textbf{AGNews}} & \multicolumn{2}{c|}{\textbf{DBPedia}} & \multicolumn{2}{c|}{\textbf{TripAdvisor}} & \multicolumn{2}{c}{\textbf{Yelp}} \\
\textbf{Model} & PPL & ACC & PPL & ACC & PPL & ACC & PPL & ACC \\ \hline
Unadapted & 96.2 & -- & 44.1 & -- & 51.6 & -- & 67.1 & -- \\
SoftmaxBias & 95.1 & \textbf{90.6} & 40.4 & 95.5 & 48.8 & 51.9 & 66.9 & 51.6 \\
ConcatCell & 93.8 & 89.7 & 39.5 & 97.8 & 48.3 & 56.0 & 66.8 & 56.9 \\
FactorCell & \textbf{92.3} & \textbf{90.6} & \textbf{37.7} & \textbf{98.2} & \textbf{48.2} & \textbf{58.2} & \textbf{66.2} & \textbf{58.8} \\
\end{tabular}
\caption{Perplexity and classification accuracy on the test set for the four word-based datasets.}
\label{table:word_accuracies}
\end{table*}

Perplexities and classification accuracies for the four word-based datasets are presented in Table \ref{table:word_accuracies}. In each of the four datasets, the FactorCell model gives the best perplexity. For classification accuracy, there is a bigger difference between the models, and the FactorCell model is the most accurate on three out of four datasets and tied with the SoftmaxBias model on AgNews. For DBPedia and TripAdvisor, most of the improvement in perplexity relative to the unadapted case is achieved by the SoftmaxBias model with smaller relative improvements coming from the increased power of the ConcatCell and FactorCell models. For Yelp, the perplexity improvements are small; the FactorCell model is just 1.3\% better than the unadapted model.

From \newcite{Yogatama2017GenerativeAD}, we see that for AGNews, much more so than for other datasets, the unigram statistics capture the discriminating information, and it is the only dataset in that work where a naive Bayes classifier is competitive with the generative LSTM for the full range of training data. The fact that the SoftmaxBias model gets the same accuracy as the FactorCell model 
on this task suggests that topic classification tasks may benefit less from adapting the recurrent layer.

For the DBPedia and Yelp datasets, the FactorCell model beats previously reported classification accuracies for generative models \cite{Yogatama2017GenerativeAD}. However, it is not competitive with state-of-the-art discriminative models on these tasks with the full training set.  With less training data, it probably would be, based on the results in \cite{Yogatama2017GenerativeAD}. 

\begin{figure}[h]
\centering
\includegraphics[width=0.45\textwidth]{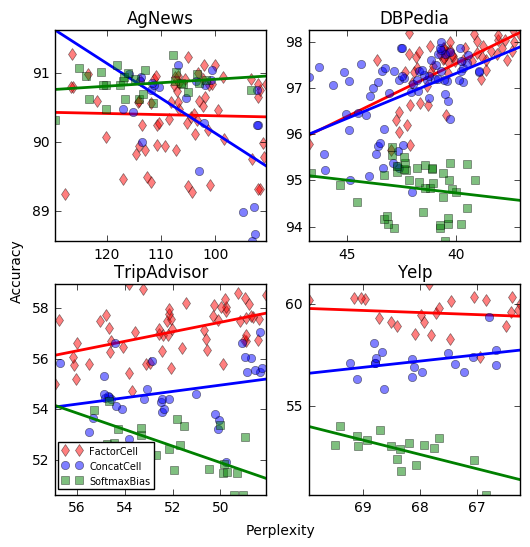}
\caption{Accuracy vs. perplexity for different classes of models on the four word-based datasets.}
\label{fig:acc_vs_ppl4}
\end{figure}

The numbers in Table \ref{table:word_accuracies} do not adequately convey the fact that there are hyperparameters whose effect on perplexity is greater than the sometimes small relative differences between models. Even the seed for the random weight initialization can have a ``major impact'' on the final performance of an LSTM \cite{reimers2017reporting}. We use Figure \ref{fig:acc_vs_ppl4} to show how the three classes of models perform across a range of hyperparameters. The figure compares perplexity on the x-axis with accuracy on the y-axis with both metrics computed on the development set. Each point in this figure represents a different instance of the model trained with random hyperparameter settings and the best results are in the upper right corner of each plot. The color/shape differences of the points correspond to the three classes of models: FactorCell, ConcatCell, and SoftmaxBias. 

Within the same model class but across different hyperparameter settings, there is much more variation in perplexity than in accuracy. The LSTM cell size is mainly responsible for this; it has a much bigger impact on perplexity than on accuracy. It is also apparent that the models with the lowest perplexity are not always the ones with the highest accuracy. See Section \ref{sec:hyperparam_analysis} for further analysis.

\begin{figure}
\centering
SoftmaxBias
\includegraphics[width=0.45\textwidth]{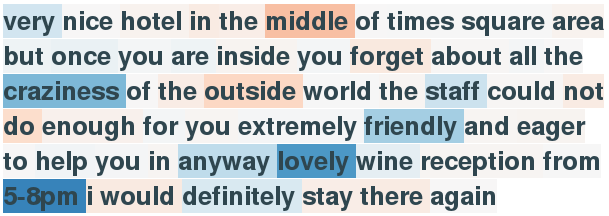}
ConcatCell
\includegraphics[width=0.45\textwidth]{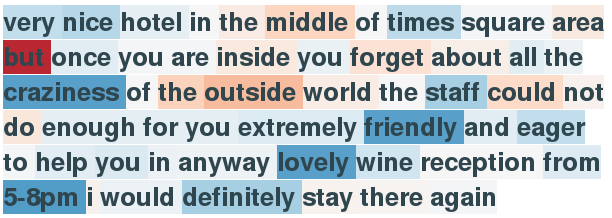}
FactorCell
\includegraphics[width=0.45\textwidth]{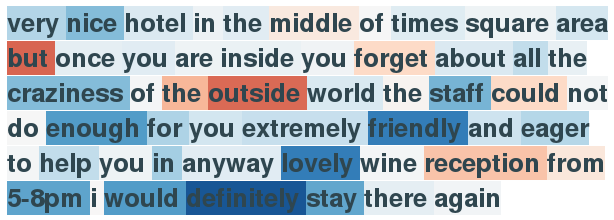}
\caption{Log likelihood ratio between a model that assumes a 5 star review and the same model that assumes a 1 star review. Blue indicates a higher 5 star likelihood and red is a higher likelihood for the 1 star condition.}
\label{table:heatmap}
\end{figure}

Figure \ref{table:heatmap} is a visualization of the per-word log likelihood ratios between a model assuming a 5 star review and the same model assuming a 1 star review. Likelihoods were computed using an ensemble of three models to reduce variance. The analysis is repeated for each class of model. Words highlighted in blue are given a higher likelihood under the 5 star assumption.

Unigrams with strong sentiment such as ``lovely" and ``friendly" are well-represented by all three models. The reader may not consider the tokens ``craziness" or ``5-8pm" to be strong indicators of a positive review but the way they are used in this review is representative of how they are typically used across the corpus. 

As expected, the ConcatCell and FactorCell model capture the sentiment of multi-token phrases. As an example, the unigram ``enough" is 3\% more likely to occur in a 5 star review than in a 1 star review. However, ``do enough" is 30 times more likely to appear in a 5 star review than in a 1 star review. In this example, the FactorCell model does a better job of handling the word ``enough."

\subsection{Character-based Models}

Next, we evaluate the EuroTwitter and GeoTwitter models using both perplexity and a classification task. For EuroTwitter, the classification task is to identify the language. With GeoTwitter, it is less obvious what the classification task should be because the context values are continuous and not categorical. We selected six cities and then assigned each sentence the label of the closest city in that list while still retaining the exact coordinates of the Tweet. There are two cities from each country: Manchester, London, Madrid, Barcelona, New York City, and Los Angeles. Tweets from locations further than 300 km from the nearest city in the list were discarded when evaluating the classification accuracy.

Perplexities and classification accuracies are presented in Table \ref{table:twitter_results}. The FactorCell model has the lowest perplexity and the highest accuracy for both datasets. Again, the FactorCell model clearly improves on the ConcatCell as measured by classification accuracy. Consistent with our hypothesis, adapting the softmax bias is not effective for these small vocabulary character-based tasks. The SoftmaxBias model has small perplexity improvements ($<1\%$) and low classification accuracies.

\begin{table}[ht]
\centering
\begin{tabular}{c|rr|rr}
  & \multicolumn{2}{c|}{\textbf{EuroTwitter}} & \multicolumn{2}{c}{\textbf{GeoTwitter}} \\
\textbf{Model} & \multicolumn{1}{c}{PPL} & \multicolumn{1}{c|}{ACC} & \multicolumn{1}{c}{PPL} & \multicolumn{1}{c}{ACC} \\ \hline
Unadapted & 6.35 & -- & 4.64 & -- \\
SoftmaxBias & 6.29 & 43.0 & 4.63 & 29.9 \\
ConcatCell & 6.17 & 91.5 & 4.54 & 42.2 \\
FactorCell & \textbf{6.07} & \textbf{93.3} & \textbf{4.52} & \textbf{63.5} \\
\end{tabular}
\caption{Perplexity and classification accuracies for the EuroTwitter and GeoTwitter datasets.}
\label{table:twitter_results}
\end{table}

\begin{figure}[ht]
\centering
\includegraphics[width=0.45\textwidth]{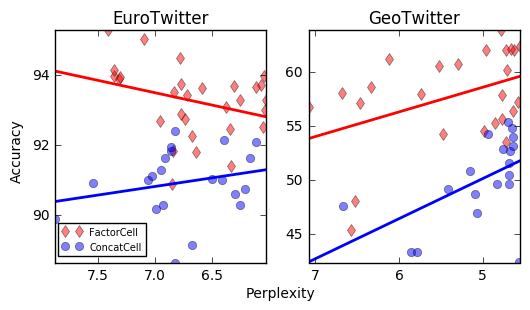}
\caption{Accuracy vs. Perplexity for different classes of models on the two character-based datasets.}
\label{fig:acc_vs_ppl2}
\end{figure}

Figure \ref{fig:acc_vs_ppl2} compares perplexity and classification accuracy for different hyperparameter settings of the character-based models. Again, we see that it is possible to trade-off some perplexity for gains in classification accuracy. For EuroTwitter, if tuning is done on accuracy rather than perplexity then the accuracy of the best model is as high as 95\%.

Sometimes there can be little to no perplexity improvement between the unadapted model and the FactorCell model. This can be explained if the provided context variables are mostly redundant given the previous tokens in the sequence. To investigate this further, we trained a logistic regression classifier to predict the language using the state from the LSTM at the last time step on the unadapted model as a feature vector. Using just 30 labeled examples per class it is possible to get 74.6\% accuracy. Furthermore, we find that a single dimension in the hidden state of the unadapted model is often enough to distinguish between different languages even though the model was not given any supervision signal. This finding is consistent with previous work that showed that individual dimensions of LSTM hidden states can be strong indicators of concepts like sentiment \cite{karpathy2015visualizing,radford2017}. 

\begin{figure}[ht]
\centering
\includegraphics[width=0.47\textwidth]{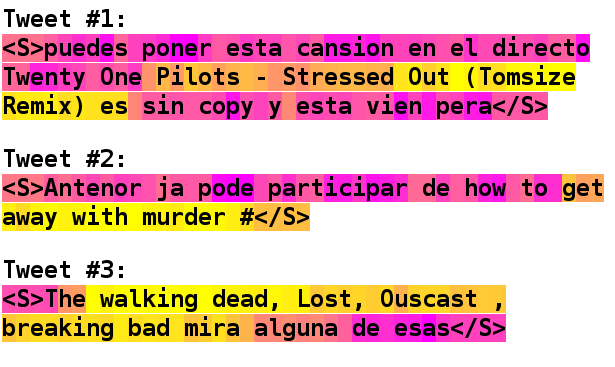}
\caption{The value of the dimension of the LSTM hidden state in an unadapted model that is the strongest indicator for Spanish text for three different code-switched Tweets.}
\label{fig:heatmap}
\end{figure}

Figure \ref{fig:heatmap} visualizes the value of the dimension of the hidden layer that is the strongest indicator of Spanish on three different code-switched tweets. Code-switching is not a part of the training data but it provides a compelling visualization of the ability of the unsupervised model to quickly recognize the language. The fact that it is so easy for the unadapted model to pick-up on the identity of the contextual variable fits with our explanation for the small relative gain in perplexity from the adapted models in these two tasks.

\subsection{Hyperparameter Analysis}
\label{sec:hyperparam_analysis}

The hyperparameter with the strongest effect on perplexity is the size of the LSTM. This was consistent across all six datasets. The effect on classification accuracy of increasing the LSTM size was mixed. Increasing the context embedding size generally helped with accuracy on all datasets, but it had a more neutral effect on TripAdvisor and Yelp and increased perplexity on the two character-based datasets. For the FactorCell model, increasing the rank of the adaptation matrix tended to lead to increased classification accuracy on all datasets and seemed to help with perplexity on AGNews, DBPedia, and TripAdvisor. 

\begin{figure}[h]
\centering
\includegraphics[width=0.47\textwidth]{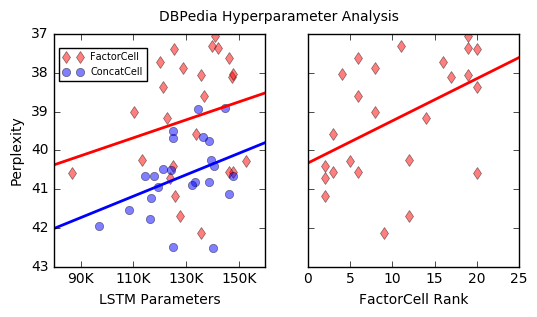}
\caption{Comparison of the effect of LSTM parameter count and FactorCell rank hyperparameters on perplexity for DBPedia.}
\label{fig:rank}
\end{figure}

Figure \ref{fig:rank} compares the effect on perplexity of the LSTM parameter count and the FactorCell rank hyperparameters. Each point in those plots represents a separate instance of the model with varied hyperparameters. In the right subplot of Figure \ref{fig:rank}, we see that increasing the rank hyperparameter improves perplexity. This is consistent with our hypothesis that increasing the rank can let the model adapt more. The variance is large because differences in other hyperparameters (such as hidden state size) also have an impact.

In the left subplot we compare the performance of the FactorCell with the ConcatCell as the size of the word embeddings and recurrent state change. The x-axis is the size of the $\mathbf{W}$ recurrent weight matrix, specifically $3 (e + d) d$ for an LSTM with $3$ gates. Since the adapted weights can be precomputed, the computational cost is roughly the same for points with the same x-value. For a fixed-size hidden state, the FactorCell model has a better perplexity than the ConcatCell. 

Since performance can be improved both by increasing the recurrent state dimension and/or by increasing rank, we examined the relative benefits of each. The perplexity of a FactorCell model with an LSTM size of 120K will improve by 5\% when the rank is increased from 0 to 20. To get the same decrease in perplexity by changing the size of the hidden state would require 160K parameters, resulting in a significant computational advantage for the FactorCell model.

Using a one-hot vector for adapting the softmax bias layer in place of the context embedding when adapting the softmax bias vector tended to have a large positive effect on accuracy leaving perplexity mostly unchanged. Recall from Section \ref{sec:softmaxbias} that if the number of values that a context variable can take on is small then we can allow the model to choose between using the low-dimensional context embedding or a one-hot vector. This option is not available for the TripAdvisor and the GeoTwitter datasets because the dimensionality of their one-hot vectors would be too large. The method of adapting the softmax bias is the main explanation for why some ConcatCell models performed significantly above/below the trendline for DBPedia in Figure \mbox{\ref{fig:acc_vs_ppl4}}.

We experimented with an additional hyperparameter on the Yelp dataset, namely the inclusion of layer normalization \cite{ba2016layer}. (We had ruled-out using layer normalization in preliminary work on the AGNews data before we understood that AGNews is not representative, so only one task was explored here.) Layer normalization significantly helped the perplexity on Yelp ($\approx 2\%$ relative improvement) and all of the top-performing models on the held-out development data had it enabled.

\subsection{Analysis for Sparse Contexts}
\label{sec:TA_anal}

The TripAdvisor data is an interesting case because the original context space is high dimensional (3500 hotels $\times$ 5 user ratings) and sparse. Since the model applies end-to-end learning, we can investigate what the context embeddings learn. In particular, we looked at location (hotels are from 25 cities in the United States) and class of hotel, neither of which are input to the model. All of what it learns about these concepts come from extracting information from the text of the reviews.

\begin{figure}
    \centering
    \includegraphics[width=0.35\textwidth]{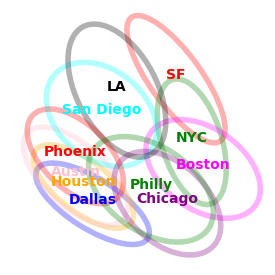}
    \caption{Distribution of a PCA projection of hotel embeddings PCA from the TripAdvisor FactorCell model showing the grouping of the hotels by city.}
    \label{fig:location_pca}
\end{figure}

To visualize the embedding, we used a 2-dimensional PCA projection of the embeddings of the 3500 hotels.
We found that the model learns to group the hotels based on geographic region; the projected embeddings for the largest cities are shown in Figure \ref{fig:location_pca}, plotting the $1.5\sigma$ ellipsoid of the Gaussian distribution of the points. (Actual points are not shown to avoid clutter.) Not only are hotels from the same city grouped together, cities that are close geographically appear close to each other in the embedding space. Cities in the Southwest appear on the left of the figure, the West coast is on top and the East coast and Midwest is on the right side. 
This is likely due in part to the impact of the region on activities that guests may mention, but there also appears to be a geographic sampling bias in the hotel class that may impact language use.

Class is a rating from an independent agency that indicates the level of service and amenities that customers can expect to receive at a hotel. Whereas, the star rating is the average score given to each establishment by the customers who reviewed it. Hotel class does not determine star rating although they are correlated ($r=0.54$). 
The dataset does not contain a uniform sample of hotel classes from each city. 
The hotels included from Boston, Chicago, and Philly are almost exclusively high class and the ones from L.A. and San Diego happen to be low class, so the embedding distributions also reflect hotel class: lower class hotels towards the top left and higher class hotels towards the bottom right. The visualization for the ConcatCell and SoftmaxBias models are similar.

Another way of understanding what the context embeddings represent is to compute the softmax bias projection $\mathbf{Q}c$ and examine the words that experience the biggest increase in probability. We show three examples in Table \ref{table:boosted}. In each case, the top words are strongly related to geography and include names of neighborhoods, local attractions, and other hotels in the same city. The top boosted words are relatively unaffected by changing the rating. (Recall that the hotel identifier and the user rating are the only two inputs used to create the context embedding.) This table combined with the other visualizations indicates that location effects tend to dominate in the output layer, which may explain why the two models adapting the recurrent network seem to have a bigger impact on classification performance.

\begin{table*}[]
\begin{tabular}{ccccp{7.5cm}}
\textbf{Hotel}       & \textbf{City} & \textbf{Class} & \textbf{Rating} & \textbf{Top Boosted Words}    \\ \hline
Amalfi               & Chicago       & 4.0            & 5               & amalfi, chicago, allegro, burnham, sable, michigan, acme, conrad, talbott, wrigley   \\
BLVD Hotel Suites    & Los Angeles   & 2.5            & 3               & hollywood, kodak, highland, universal, reseda, griffith, grauman's, beverly, ventura \\
Four Points Sheraton & Seattle       & 3.0            & 1               & seattle, pike, watertown, deca, needle, pikes, pike's monorail, uw, safeco          
\end{tabular}
\centering
\caption{The top boosted words in the Softmax bias layer for different context settings in a FactorCell model.}
\label{table:boosted}
\end{table*}

\section{Prior Work}
\label{sec:prior}

There have been many studies of neural language models that can be dynamically adapted based on context.  Methods have been referred to as context-dependent models \cite{mikolov2012context}, context-aware models \cite{TangContextAware},  conditioned models \cite{Ficler2017ControllingLS}, and controllable text generation \cite{Hu2017ControllableTG}. These models have been used in scoring word sequences (such as for speech recognition or machine translation), for text classification, and for generation. In some work, context corresponds to the previous word history. Here, we instead consider known factors such as user, location and domain metadata, though the framework could be used with history-based context.

The studies that most directly relate to our work are neural models that correspond to special cases of the more general FactorCell model, including those that leverage what we call the SoftmaxBias model \cite{dieng2016topicrnn,TangContextAware,Yogatama2017GenerativeAD,Ficler2017ControllingLS} and others that use the ConcatCell approach \cite{mikolov2012context,wen2013recurrent,chen2015recurrent,ghosh2016contextual}. One study \cite{Ji2015DocumentCL} compares the two approaches, which they refer to as ccDCLM and coDCLM. They find that both approaches give similar perplexities,
but their ConcatCell style model does better at an auxiliary sentence ordering task. This is consistent with our finding that adapting at the recurrent layer can benefit certain tasks while having only a minor impact on perplexity. They do not test any models that adapt both the recurrent and output layers.
\newcite{Hoang2016IncorporatingSI} also consider adapting at the hidden layer vs.\ at the softmax layer, but their architecture does not fit cleanly into the framework of the SoftmaxBias model because they use an extra perceptron layer; thus, it is difficult to compare the experimental findings with ours.

The FactorCell model is distinguished by having an additive (factored) context-dependent transformation of the recurrent layer weight matrix. A related additive context-dependent transformation has been proposed for log-bilinear sequence models \cite{Eisenstein+11,Hutchinson+15}, but these are less powerful than the RNN. A somewhat different use of low-rank factorization has previously been used to reduce the parameter count in an LSTM LM \cite{Kuchaiev2017FactorizationTF}, finding that the reduced number of parameters leads to faster training.

There is a long history of adapting n-gram language models. (See \newcite{demori1999language} or \newcite{bellegarda2004statistical} for a survey.) One recent example is \newcite{chelba2015sparse} where a 34\% relative improvement in perplexity was obtained when using geographic features for adaptation. We hypothesize that the impressive improvement in perplexity is possible because the language in their dataset of Google mobile search queries is particularly sensitive to location. Compared to n-gram based LMs, our model has two advantages in the way that it handles context. First, as we showed in our GeoTwitter experiments, we can adapt to geography using GPS coordinates as input without using predefined geographic regions as in Chelba and Shazeer. Second, our model supports the joint effect of multiple contextual variables. Neural models have an advantage over discrete models as the number of context variables increases.

Much of the work on context-adaptive neural language models has focused on incorporating document or topic information \cite{mikolov2012context,Ji2015DocumentCL,ghosh2016contextual,dieng2016topicrnn}, 
where context is defined in terms of word or n-gram statistics. Our work differs from these studies in that the context is defined by a variety of sources, including discrete and/or continuous metadata, which is mapped to a context vector in end-to-end training.
Context-sensitive language models for text generation tend to involve other forms of context similar to the objective of our work, including speaker characteristics \mbox{\cite{LuanRole,li2016persona}}, dialog act \cite{WenCU2015}, sentiment and other factors \mbox{\cite{TangContextAware,Hu2017ControllableTG}}, and style \cite{Ficler2017ControllingLS}.
As noted earlier, some of this work has used discriminative text classification to evaluate generation. In preliminary experiments with the Yelp data set, we found that the generative classifier accuracy of our model is highly correlated with discriminative classfier accuracy (\mbox{$r \approx~0.95$}). Thus, by this measure, we anticipate that the model would be useful for generation applications. Anecdotally, we find that the model gives more coherent generation results for DBPedia data, but further validation with human ratings is necessary to confirm the benefits on more sources.

\section{Conclusions}
\label{sec:concl}

In summary, this paper has introduced a new model for adapting (or controlling) a language model depending on contextual metadata. The FactorCell model extends prior work with context-dependent RNNs by using the context vector to generate a low-rank, factored, additive transformation of the recurrent cell weight matrix. Experiments with six tasks show that the FactorCell model matches or exceeds performance of alternative methods in both perplexity and text classification accuracy. Findings hold for a variety of types of context, including high-dimensional contexts, and the adaptation of the recurrent layer is particularly important for character-level models. For many contexts, the benefit of the FactorCell model comes with essentially no additional computational cost at test time, since the transformations can be pre-computed. 
Analyses of a dataset with a high-dimensional sparse context vector show that the model learns context similarities to facilitate parameter sharing.

In all six tasks that are explored here, all context factors are available for all training and testing samples. In some scenarios, it may be possible for some context factors to be missing. A simple solution for handling this is to use the expected value for the missing variable(s), since this is equivalent to using a weighted combination of the adaptation matrices for the different possible values of the missing variables.

In this work, the experiment scenarios all used metadata to specify context, since this type of context can be more sensitive to data sparsity and has been less studied. In contrast, in many prior studies of language model adaptation, context is specified in terms of text samples, such as prior user queries, prior sentences in a dialog, other documents related in terms of topic or style, etc. The FactorCell framework introduced here is also applicable to this type of context, but the best encoding of the text into an embedding (e.g. using bag of words, sequence models, etc.) is likely to vary with the application. 
The FactorCell can also be used with online learning of context vectors, e.g.\ to take advantage of previous text from a particular author (or speaker) \cite{jaech2018}.

The models evaluated here were tuned to minimize perplexity, as is typical for language modeling. In analyses of performance with different hyperparameter settings, we find that perplexity is not always positively correlated with accuracy, but the criteria are more often correlated for approaches that adapt the recurrent layer. While not surprising, the results raise concerns about using perplexity as the sole evaluation metric for context-aware language models. More work is needed to understand the relative utility of these objectives for language model design.

\bibliography{mybib}
\bibliographystyle{acl2012}

\appendix
\section{LSTM FactorCell Equations}
\label{sec:lstm_equations}

Only trivial changes are needed to use the FactorCell method on an LSTM instead of a vanilla RNN. Here, we list the equations for an LSTM with coupled input and forget gates, which is what was used in our experiments. 

The weight matrix $\mathbf{W}$ from Equation \ref{eq:rnn} is now size 
$3d \times (e+d)$ and $b$ is dimension $3d$,
where $3$ is the number of gates. Likewise, $\mathbf{Z}_R$ from Equation \ref{eq:w_prime} is made to be of size $r \times 3d \times k$. The weight matrix $\mathbf{W}'$ is as defined in Equation \ref{eq:FactorCell} and after computing it's product with the input $[w_t, h_{t-1}]$, the result is split into three vectors of equal size: $i_t$, $f_t$, and $o_t$
\begin{equation}
   [ i_t, f_t, o_t ] = \mathbf{W'}[w_t, h_{t-1}] + b ,
   \label{eq:lstm-gates}
\end{equation}
where $i_t$, $f_t$ and $o_t$ are used in the input gate, the forget gate, and the output gate, respectively.

Using these three vectors we perform the gating operations to compute $h_t$ using the memory cell $m_t$ as follows:
\begin{align}
\begin{split}
    f_t & \leftarrow \mathrm{sigmoid}(f_t + 1.0) \\
   m_t &= m_{t-1} \odot f_t + (1.0 - f_t) \odot \mathrm{tanh}(i_t) \\
    h_t &= \mathrm{tanh}(m_t) \odot \mathrm{sigmoid}(o_t)
\end{split}
\end{align}

Note that equation~\ref{eq:ConcatCell}, which shows that a context vector concatenated with input is equivalent to an additive bias term, extends to equation~\ref{eq:lstm-gates}. In other words, in the LSTM version of the ConcatCell model, the context vector effectively introduces an extra bias term for each of the 3 gates.

\end{document}